# Virtual Reality based Digital Twin System for remote laboratories and online practical Learning

CLAIRE PALMER[a,b], Ben ROULLIER[a,b], Muhammad AAMIR [a,b], Leonardo STELLA[a], Uchenna DIALA[a], Ashiq ANJUM[c], Frank MCQUADE[b], Keith COX[b] and Alex CALVERT[b]
[a] *University of Derby, Kedleston Road, Derby, DE22 1GB*
[b] *Bloc Digital, 2nd Floor, Enterprise Centre, Bridge St, Derby, DE1 3LD, UK*
[c] *University of Leicester, University Road, Leicester, LE1 7RH*

**Abstract.** There is a need for remote learning and virtual learning applications such as virtual reality (VR) and tablet-based solutions which the current pandemic has demonstrated. Creating complex learning scenarios by developers is highly time-consuming and can take over a year. There is a need to provide a simple method to enable lecturers to create their own content for their laboratory tutorials. Research is currently being undertaken into developing generic models to enable the semi-automatic creation of a virtual learning application. A case study describing the creation of a virtual learning application for an electrical laboratory tutorial is presented.

## 1. Introduction

The current pandemic crisis has demonstrated the need for remote learning and virtual learning applications such as virtual reality (VR) and tablet-based solutions. It is widely acknowledged that online learning is here to stay [1]. The Harvard Business Review states that "This moment is likely to be remembered as a critical turning point between the "time before," when analog on-campus degree-focused learning was the default, to the "time after," when digital, online, career-focused learning became the fulcrum of competition between institutions" [2]. Good quality career-focused learning requires university science and technology courses to offer practical experience replicating industrial equipment and scale. Even in normal times laboratory provision is expensive and may not supply the range, complexity or realistic indication of the dimensions encountered in industry. Due to limited provision and health and safety requirements student access to laboratories may be limited, providing an educational barrier to groups such as part-time and commuting students. In the UK there is a move towards more students commuting to campus to save money [3].

VR learning offers a safe and realistic environment with wider accessibility than physical laboratories. It is able to supply a wide range of experience by enabling users to train on multiple makes and models of equipment. VR learning can accommodate individual differences in terms of learning style [4] and students are able to repeat experiments in order to gain proficiency.

A VR learning solution requires the development of a virtual environment, e.g., a laboratory, containing a VR replica of the workshop/laboratory, equipment, equipment layout and attribute data, the laboratory procedures to be followed, and the methods enabling the user to interact with the VR learning solution. As of now, there are a variety of methods via which a user can interact with a virtual environment, e.g., mouse and keyboard, hand-held controllers as used in games consoles and multi-touch touchscreen, typically an iPad or slate. Platforms such as the Oculus Quest and the Microsoft Hololens2 offer the user the ability to engage with the virtual world in 3D and it is expected that the advent of Apple and Google XR glasses will render this commonplace. However, the prevalence of PC, Mac and tablet devices dictate that a solution must also offer a means to engage with a virtual world via a 2D interface, i.e., the screen.

Currently, building virtual learning applications requires gathering material from a wide range of formats, for example documents (such as laboratory manuals), interviews with experts and photographs of the equipment and environment. The virtual solution is then built after manually extracting the information from this material and transcribing it into software code. Creating complex learning scenarios via existing methods can take over a year. Any updates or corrections require the software to be recoded, issued as a new application and redeployed to the end user. This takes a considerable amount of developer time and is therefore expensive. Due to the number of laboratory simulations required there is a need for lecturers to create their own content for their laboratory tutorials. However, lecturers are not software developers and lack the time and resources to develop virtual applications.

Research is currently being undertaken into developing a digital twin to enable the semi-automatic creation of a virtual learning application. A modelling approach to capture laboratory learning scenarios is considered. Equipment behavior and procedural information can be modelled generically. Although each individual learning scenario is unique, employing generic models enables information to be reused across scenarios. Initial research into developing electronic training based on an underlying model has been conducted by Torres *et al*. [5]. Vergara *et al*. [6] propose a guide for designing VR learning environments in engineering, emphasizing the need for adequate realism. Barbieri *et al*. (2021) propose a methodology to develop a digital twin framework of a flow shop which generates a scheduling reactive to machine breakdown.

This paper is organized as follows. In Section 2 a digital twin architecture is presented. Section 3 describes how the architecture is employed to create a VR learning application. A case study demonstrating the approach is described in section 4. Ideas for future developments are discussed.

**2. Digital Twin**

A digital twin can be defined as a digital or virtual representation of a real or potential product, system, process or value chain, although how the term is applied throughout the product lifecycle may vary. However, its principal use is to foster a common understanding of the system in question. Digital twins can be applied in many domains, e.g. manufacturing, industrial Internet of Things, Healthcare, Smart Cities, Automobile and retail [7]. The application of the digital twin approach is limited by the complexity of creating equipment models which determines cost effectiveness. The advantages of applying a modelling approach to implement the digital twin is that the generation of equipment and operating procedures is facilitated through semi-automated creation based on generic models. The use of generic models aids re-use and removes the need for repetitious low-level coding, preventing duplication and the inconsistencies which can result from manual coding and reducing development costs.

The Digital Twin Architecture consists of three main elements: first, the Digital Twin Builder application which enables a lecturer to specify laboratory tutorials via an easy to use drag and drop interface and store these specifications in a serialised file format for re-use. Second, three processing pipelines (data, geometry, and process) which characterise incoming information and make it available within the drag and drop interface. Finally, the Digital Twin Player, which provides the VR learning application, enabling students to interact directly with the digital twin produced and to share results with fellow learners and the lecturer. The scenario definition file created with the Digital Twin Builder separates the virtual world creation from the virtual world viewer. This allows the same digital twin configuration to support multiple form factors and engage users in an optimal manner (e.g., XR glasses, tablet etc.).

The Digital Twin Architecture is shown in Figure 1. The figure shows the process of importing data, geometry and processes into the Digital Twin Builder and producing Digital Twin specifications which can be used with the Digital Twin Player(s) to view, interact with, and collaborate around.

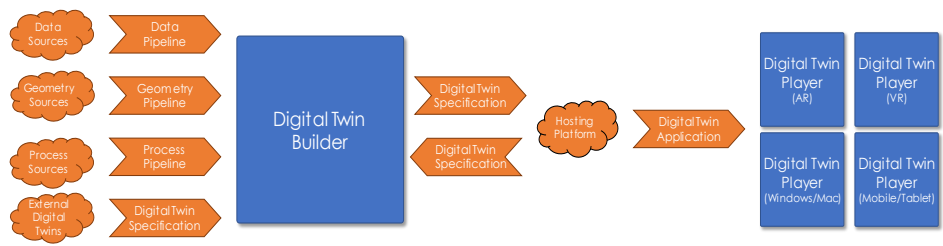

**Figure 1.** Digital Twin Builder Architecture.

*2.1. The Data Pipeline*

The VR learning application utilises data to produce simulation results. The range of data which could be included is huge, for example, pressure maps, temperature changes and airflows. The Data Pipeline imports generic data models which the lecturer links to relevant laboratory equipment objects. Data is imported within a spreadsheet and the lecturer selects columns of interest. A variable number of columns may be selected. The prototype is capable of evaluating mathematical expressions based on inputs from within a digital twin and controlling aspects of the twin based on simulation results. Data is displayed within the Digital Twin Player as a graphs, textual displays and as desired equipment behaviors.

*2.2. The Geometry Pipeline*

Geometry is required to create virtual models of laboratory equipment items, together with their sizes and positions within the virtual world. In order to use this geometry in AR/VR however, it must be represented in a format suitable for real-time rendering. This conversion has traditionally been performed by experts who must balance visual quality against rendering performance, in a time-consuming process. This precludes the direct use of CAD models (commonly used for technical, engineering and scientific equipment) in VR/AR digital twins. An intelligent pipeline has been developed which utilises machine learning techniques to automate the conversion of CAD data to digital twin geometry [9]. Three Random Forest Classifiers are incorporated within an iterative framework. Commercial tools are used to convert geometry, while a machine learning framework based on geometric measurements is used to control the behavior of these tools. Figure 2 shows the architecture of this pipeline.

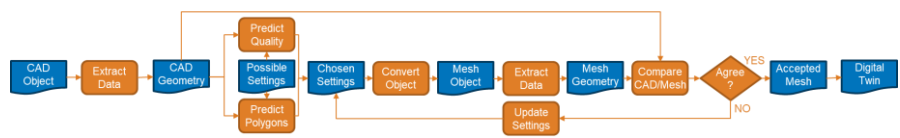

**Figure 2.** The Architecture of the Geometry Pipeline

*2.3. The Process Pipeline*

A VR learning application will contain processes to guide a student as to which tasks are needed to undertake the required assignment or tutorial. A process is defined as an ordered or unordered set of steps. Manual creation of these processes is time-consuming and error prone, requiring translation of tutorial instructions into machine readable data and linking this data to equipment objects, user inputs, and physical processes.

The Process Pipeline captures procedure information in domain specific models generated from human readable structured forms. Two types of process model step exist: procedure and instruction. Procedures contain steps (i.e., a procedure can contain procedures, instructions, or both procedures and instructions). An instruction refers to one or more equipment items within the Digital Twin, e.g., "position speedometer", "use a cable to connect DC motor to ammeter". Ordered steps contain a link to the next step in the process. Instructions contain a condition which must be satisfied in order to progress to the next step. These conditions are based on the user altering an equipment item to a defined state (such as pressing a button to "on") or external factors (wait ten seconds). The domain specific models enable hierarchical ordering of process stages, inclusion of equipment references, and multiple forms of interdependency between process stages and equipment states.

**3. Creating the VR Learning Application**

To create a VR learning application using the Digital Twin architecture the following information is needed: a list of equipment and 3-D representations of these equipment items for display in the virtual world; a virtual environment in which to place the equipment (e.g. workbench, laboratory area); the location of the equipment within the virtual world; the location of VR equipment interactions (i.e. the area of the equipment which interfaces with another item during assembly or with the user during operation) and user interaction type (i.e. whether to push a button, turn a dial, add a cable); a list of tutorial instructions; and data to enable simulation.

The Geometry Pipeline provides 3-D equipment model representations for the equipment items and virtual environment. The representations are imported into the Digital Twin Builder as reusable generic object models. The Digital Twin Builder provides a user-friendly form and a drag and drop interface to enable VR equipment interactions to be defined for the generic objects. To create a VR learning application a scenario is created within the Digital Twin Builder and a virtual environment is added to it. Object models are instantiated into the scenario and location information is added to these equipment instances via a form and drag and drop interface. The Data Pipeline and Process Pipeline enable simulation data and tutorial instructions to be added to the VR learning application.

**4. Case Study: An Electrical Laboratory Tutorial**

The brief shown in figure 3 was provided as a test case for the Digital Twin Architecture.

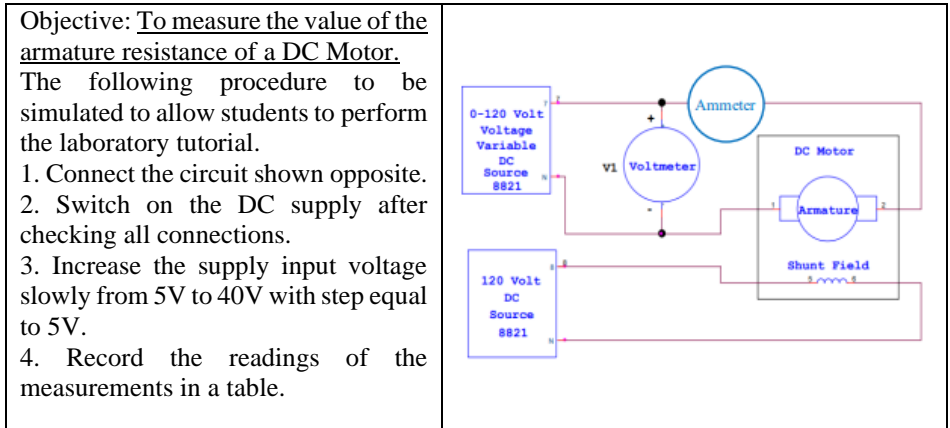

**Figure 3.** The Case Study Brief.

The Unity development platform [10] was used to implement the Digital Twin Architecture. Figure 4 shows a screen shot of the electrical laboratory tutorial being specified within the Digital Twin Builder (left) and a case study equipment item is shown within the Digital Twin Player (right). The left hand panel of the Digital Twin Builder lists the equipment instances and procedures which have been created to fulfil the case study brief. The center panel shows the virtual learning scenario created. The arrows protruding from the equipment instances indicate VR interactions. A dial interaction has been selected and its values are shown in the right hand panel of the Digital Twin Builder. The right hand panel provides an example of the forms via which equipment and process information can be created and edited. The sphere shown in the center panel constitutes the drag and drop interface which provides an alternative means to provide equipment location, size and rotation information.

The Digital Twin Player shows the DC motor with its data simulation results displayed as a graph. As the Digital Twin Player runs the DC motor rotates simultaneously in response to the values of the speed data shown.

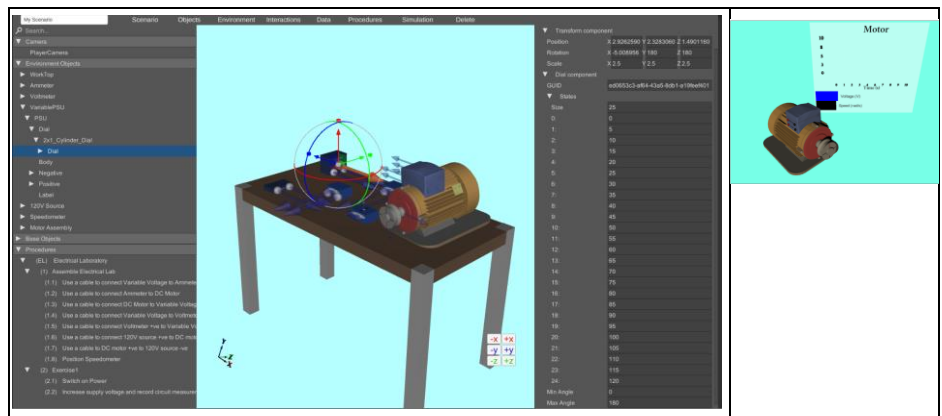

**Figure 4.** Creating the Case Study within the Digital Twin Architecture.

**5. Conclusions and Future Work**
A Digital Twin Architecture has been defined which enables a virtual learning application to be semi-automatically created. The Digital Twin Architecture utilizes a generic modelling approach, saving work for lecturers through allowing information re-use between virtual tutorials. The current version utilizes current equipment data. It is envisaged that a future version will allow students to control equipment remotely. It was discovered that whilst this new approach can shorten the time taken to create a VR learning application from weeks to hours, the most time-consuming aspect is the need to accurately position the VR interactions to enable equipment items to connect together, involving iteration between the Digital Twin Editor and Player to check location. Future work includes investigating a cloud based service for the digital twin for remote access [11,12,13,14], allowing high throughput creation of digital twins, advanced object recognition using machine learning models [15] and Edge enhanced digital twins for performance and real time analytics [16,17,18,19]. Also, further research is required into the best form of interface to create the VR interactions and into whether semi-automatic positioning is possible.

**Acknowledgements**

This work was supported by Innovate UK Knowledge Transfer Partnership under Grant agreement no.11936 and Bloc Digital.